%% file: emnlp_main.tex
\documentclass[11pt]{article}

\PassOptionsToPackage{table,dvipsnames}{xcolor}

\usepackage[final]{acl}

\usepackage{times}
\usepackage{latexsym}
\usepackage[T1]{fontenc}
\usepackage[utf8]{inputenc}
\usepackage{microtype}

\usepackage{amsmath}
\usepackage{amssymb}
\usepackage{dsfont}
\providecommand{\mathbbm}[1]{\mathds{#1}}
\usepackage{mathtools}
\usepackage{amsthm}

\usepackage{booktabs}
\usepackage{xcolor} 
\usepackage[most]{tcolorbox}
\usepackage{graphicx}
\usepackage{enumitem}
\usepackage{tabularx}
\usepackage{multirow}
\usepackage{arydshln}
\usepackage{algorithm}
\usepackage{wrapfig}
\usepackage{algpseudocode}
\usepackage{caption}
\usepackage{adjustbox}
\definecolor{MyCellShade}{HTML}{E8EBF9} 

\usepackage{breqn}            

\usepackage{xspace}

\newcommand{\statebelief}[1]{\hat{s}_{#1}}

\newcommand{\statetrue}[1]{{s}_{#1}}

\newcommand{\tokenthinks}{\texttt{<think>}}
\newcommand{\tokenthinke}{\texttt{</think>}}
\newcommand{\tokennexts}{\texttt{<prediction>}}
\newcommand{\tokennexte}{\texttt{</prediction>}}
\newcommand{\tokencurs}{\texttt{<observation>}}
\newcommand{\tokencure}{\texttt{</observation>}}

\newcommand{\tokenans}{\texttt{<answer>}}
\newcommand{\tokenane}{\texttt{</answer>}}

\definecolor{customgray}{rgb}{0.25,0.25,0.25}
\definecolor{customred}{rgb}{0.8,0.05,0.05}
\definecolor{urlcolors}{rgb}{0.872,0.2,0.552}

\tcbuselibrary{listings}
\newtcblisting{promptbox}[2][]{%
  breakable,
  colback=blue!10!gray!10!white,
  colframe=blue!20!gray,
  boxrule=0.6pt, arc=2pt,
  left=2pt,right=2pt,top=2pt,bottom=2pt,
  fonttitle=\bfseries\small, title={#2},
  listing only,
  before skip=2pt, after skip=2pt,
  listing options={
    basicstyle=\ttfamily\small,
    columns=fullflexible,
    keepspaces=true,
    breaklines=true,
    breakatwhitespace=true,
    breakautoindent=false,
    breakindent=0pt,
    postbreak=,
    prebreak=,
    showstringspaces=false,
    xleftmargin=0pt,
    framexleftmargin=0pt,
    escapeinside={(*@}{@*)},
    literate=
      {√}{{$\surd$}}1
      {✓}{{$\checkmark$}}1
  },
  #1}

\newtcblisting{promptplain}{%
  enhanced, sharp corners, blanker,
  borderline={0.25pt}{0pt}{black},
  listing only,
  listing options={
    basicstyle=\ttfamily\small,
    breaklines=true,
    columns=fullflexible,
    keepspaces=true,
    showstringspaces=false,
  },
  left=6pt,right=6pt,top=6pt,bottom=6pt,
  before skip=10pt, after skip=10pt,
}

\newcommand{\up}[1]{\textcolor{OliveGreen}{\tiny \ $\uparrow${#1}}}
\newcommand{\downbad}[1]{\textcolor{Maroon}{\tiny \ $\downarrow${#1}}}

\colorlet{cameracolor}{black}
\newcommand{\camera}[1]{{\color{cameracolor}#1}}
\newenvironment{camerablock}{\par\color{cameracolor}}{\par}

\definecolor{darkblue}{rgb}{0, 0, 0.5}
\hypersetup{colorlinks=true, citecolor=darkblue, linkcolor=darkblue, urlcolor=darkblue}

\title{Why Do LLM Agents Fail in Exploring New Environments? A World-Modeling Perspective}

\author{%
  Shiqi Chen\thanks{Corresponding Author}\,\thanks{Core Contributor}\,$^{1}$, %
  Tongyao Zhu\footnotemark[2]\,$^{7}$, %
  Zian Wang\footnotemark[2]\,$^{8}$, %
  Jinghan Zhang\footnotemark[2]\,$^{3}$, %
  Kangrui Wang$^{2}$, \\
  \bfseries Ruochen Zhou$^{1}$, %
  Siyang Gao$^{1}$, %
  Teng Xiao\footnotemark[2]\,$^{5,6}$, %
  Yee Whye Teh$^{4}$, %
  Junxian He\footnotemark[1]\,$^{3}$, %
  Manling Li\footnotemark[1]\,$^{2}$ \\[3pt]
  {\normalfont\small $^{1}$City University of Hong Kong, $^{2}$Northwestern University} \\
  {\normalfont\small $^{3}$The Hong Kong University of Science and Technology, $^{4}$Oxford University} \\
  {\normalfont\small $^{5}$Allen Institute for AI (AI2), $^{6}$University of Washington, $^{7}$National University of Singapore} \\
  {\normalfont\small $^{8}$The Hong Kong Polytechnic University}%
}

\begin{document}
\maketitle

\begin{abstract}
\input{sections/1-abs}
\end{abstract}

\section{Introduction}

\input{sections/2-intronew}

\section{Unseen Environments Induce Exploration Collapse}
\input{sections/3-collapse}

\section{SPA: Internal World Modeling Restores Exploration}
\input{sections/3-method}

\section{Explore-then-exploit outperforms reward shaping for world modeling}
\input{sections/4-exp}

\section{What makes the Explore-then-Exploit World Modeling work?}
\label{sec:ablation}
\input{sections/5-ablation}
\input{sections/6-findings}

\section{\texorpdfstring{\camera{Related Work}}{Related Work}}
\label{sec:relwork}
\input{sections/7-relwork}

\input{sections/8-conclusion}

\section*{Limitations}
\paragraph{Scope.} Our main results are on grid-style interactive benchmarks
(Sokoban, FrozenLake, Sudoku) with small- to mid-scale open-weight models;
ALFWorld provides only a single-model check. Whether the same recipe transfers
to larger models and richer, real-world agentic tasks (e.g., web navigation,
tool use, embodied control) remains to be verified.

\paragraph{Reliance on ground-truth states.} \texttt{SPA} supervises the
self-experience SFT stage with ground-truth states and coordinates exposed by
the simulator, and our ablations indicate this supervision is necessary for the
gains. Environments without accessible privileged state would first require
estimating it (e.g., via a learned perception module), which we do not study
here.
\camera{We view this access as a matter of experimental scope rather than a
property of the method: environment-provided or deterministically extracted
structured states give a controlled setting for isolating whether stronger
state/transition grounding improves RL under unfamiliar states, not a claim
that privileged state is available in general deployments.}

\paragraph{Evidence for the mechanism.} We evidence restored exploration
through the recovery of Pass@k\camera{, complemented by post-hoc trajectory
metrics (successful-path coverage and action diversity;
\S\ref{sec:findings}, Appendix~\ref{appdx:analyses})}; \camera{a fully formalized account of the
exploration--exploitation tradeoff (e.g., in terms of policy entropy or
regret) and} a finer-grained
causal account of \emph{why} transition-prediction SFT restores exploration
\camera{are} left to future work. Relatedly, the internalized world model is used as a
grounded \emph{initialization} for RL rather than an inference-time planner, so
our claims concern initialization, not explicit planning.

\paragraph{Stability and comparisons.} Training remains unstable in stochastic
settings, and errors in the instruction/data-generation stage degrade the
supervised transition data. Our VAGEN comparison is a reimplementation under
matched settings and may not reflect its best configuration. Future work
includes uncertainty-aware transition modeling and scaling to richer
modalities. Due to compute constraints, we report numbers based on single runs, but we repeat the experiments across different models, environments, and ablations. Future work could investigate the stability of SPA training as well. We do not foresee that our work carries risks that are worth mentioning here.  

\begin{camerablock}
\section*{Acknowledgments}
We thank Zihan Wang, the lead author of RAGEN, for numerous insightful suggestions on this project and for sharing technical details of the RAGEN codebase. We also thank Zichen Liu for helpful suggestions on cross-game generalization evaluations, and Chang Ma, Wei Liu, Junteng Liu and Zhengyu Chen for valuable discussions. We also thank Shuaicheng Niu for pointing out some typos of the draft. YWT is supported by the Ministry of Digital Development and Information (MDDI) under the Singapore Global AI Visiting Professorship Program (Award No. AIVP-2024-002).
\end{camerablock}

\bibliography{iclr2025_conference}

\appendix

\section{Prompts}
\label{appdix: prompts}
\input{sections/appendix-prompts}
\input{sections/appendix-expsetup}

\section{Full Results}
\input{sections/appendix-fullres}

\input{sections/appendix-analyses}

\section{Use of Large Language Models (LLMs)}
\input{sections/appendix-llm}

\section{SPA on Alfworld}
\input{sections/appendix-alfworld}

\section{SPA on GRPO}
\input{sections/appendix-grpo}

\end{document}

%% file: sections/1-abs.tex
\vspace{-5pt}
Large Language Models (LLMs) as agents often fail to improve in \emph{new} environments. We identify and characterize a failure mode we call \emph{exploration collapse}: under reinforcement learning (RL) in environments whose states are unfamiliar to the policy, \emph{Pass@k}, the probability that at least one of \(k\) sampled trajectories succeeds, drops markedly over training even as \emph{Pass@1} edges up, revealing increasingly brittle exploration; environments closer to the pretraining distribution show no such decline.
We trace this collapse to weak grounding in environment states and dynamics, and study a simple remedy: explicitly teaching the agent to estimate the current state and predict its transitions before optimizing for reward. We instantiate it as \texttt{SPA}, an explore-then-exploit recipe that cold-starts the policy with a Self-Experience supervised finetuning (SFT) stage, collecting the model's own interaction trajectories and supervising state and next-state prediction, and then runs standard RL. The resulting world model serves as a grounded \emph{initialization} for RL rather than an inference-time planner.
Across unseen environments, \texttt{SPA} consistently and substantially improves over vanilla RL: for example, it raises the Sokoban success rate from \textbf{25.6\%} to \textbf{59.8\%} on Qwen2.5-1.5B-Instruct, letting sub-3B models surpass a 20B baseline on these tasks. Controlled studies indicate that the gains track four factors: grounded state representations, explicit transition modeling, self-experience trajectories from a sufficiently strong exploration policy, and adequate coverage of transition data.

%% file: sections/2-intronew.tex
\begin{figure}[t]
  \centering
  \includegraphics[width=\columnwidth]{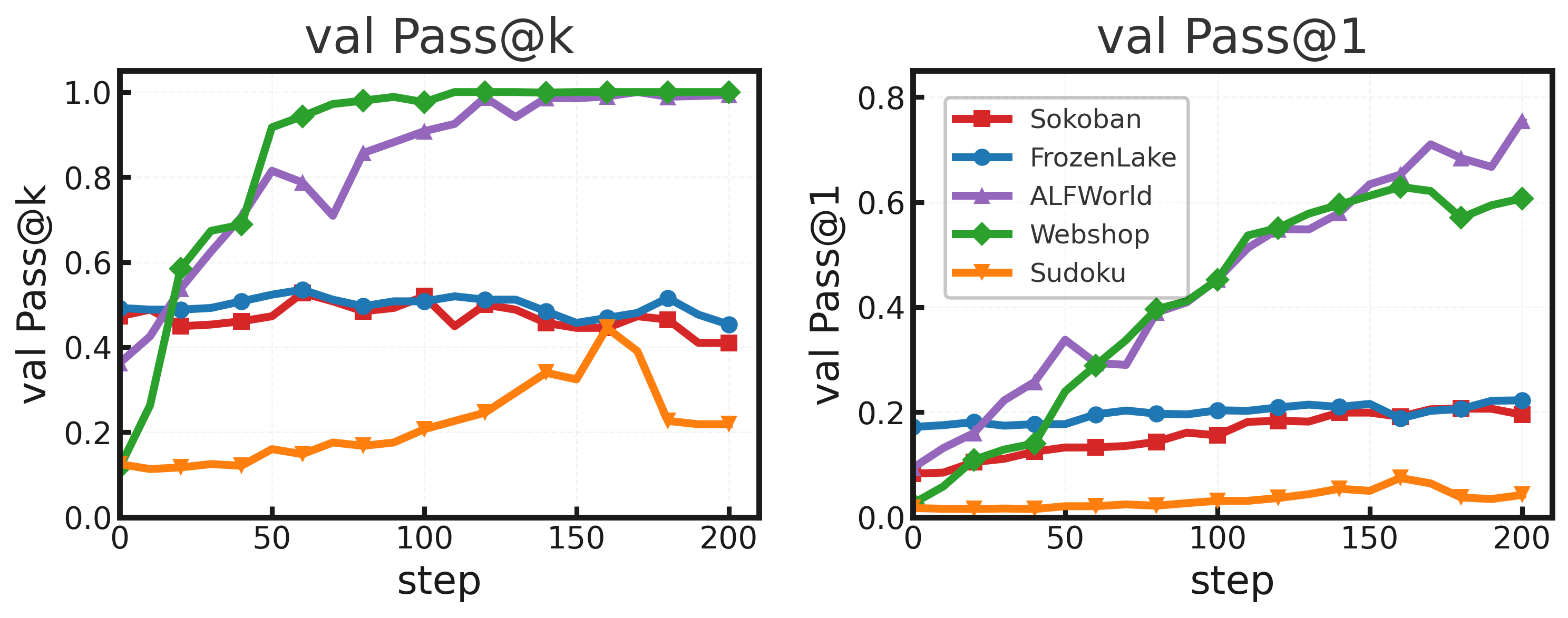}
  \caption{Validation score on five environments. Sokoban, FrozenLake and Sudoku are OOD for the models, where \emph{Pass@k} ($k$=8) drops with training. However, in ALFWorld and WebShop, \emph{Pass@k} increases greatly with training.}
  \label{fig:teaser}
\end{figure}
Agentic reinforcement learning (RL) has become a primary framework for finetuning Large Language Model (LLM) agents~\citep{wang2025ragen,jin2025search}. Using ReAct-style scaffolds~\citep{yao2023react} and optimizing for task rewards, this approach has been applied to computer-use planning~\citep{cao2025large}, tool use~\citep{schick2023toolformer}, and web search~\citep{wu2025webdancer,jin2025search}.
However, the performance degrades sharply in unseen environments that are beyond the LLM's pre-training distribution. In such cases, the encountered state is unfamiliar to the policy model. We quantify this mismatch by combining the perplexity (PPL) of the state description with that of the state space. As shown in Table~\ref{tab:state-ppl}, Sokoban, FrozenLake, and Sudoku exhibit state PPLs far above random guess under these environments, yet ALFWorld and WebShop yield much lower state PPLs compared with random guess, indicating a stronger distributional shift in the former environments.

To diagnose when LLM agents struggle to learn in unseen environments, 
we empirically investigate agent performance during RL fine-tuning and uncover a systematic divergence between two key metrics: \textit{Pass@1}, the success rate of the highest-probability trajectory, and \textit{Pass@k}, whether at least one of the $k$ sampled trajectories is successful. Pass@k further captures whether an agent leverages a learned world model to diversify attempts within a small rollout budget, without reward overfitting or brittle credit assignment.
As shown in Figure~\ref{fig:teaser}, in unseen environments like Sokoban, FrozenLake, and Sudoku, standard RL causes Pass@k performance to consistently degrade, while Pass@1 increases marginally. This is different from familiar settings (e.g., ALFWorld and Webshop), where both metrics improve. 
This divergence reveals a fundamental limitation: current agentic RL develops a narrow, exploitative solution path 
(reflected in {Pass@1}), but fail to explore across multiple paths (in {Pass@k}), a phenomenon we call \textbf{exploration collapse}.
\camera{We note that related large-$k$ narrowing has been observed for single-turn RL with verifiable rewards~\citep{yue2025does,wen2025reinforcement,zhai2026does}; our contribution is to study its manifestation in unfamiliar \emph{multi-turn} environments, diagnose it as a grounding failure, and mitigate it with a grounded initialization (\S\ref{sec:relwork}).}
To this end, we ask an important research question: \textbf{How can we enable effective exploration for scalable agentic reinforcement learning in unseen environments?}
We argue that a key lever for restoring exploration in unseen environments is to \emph{ground} the agent in the environment's states and dynamics---internalizing what we refer to as a \textbf{world model}~\citep{xing2025critiquesworldmodels} of its underlying rules.
Unlike classical non-LLM RL that queries an explicit world model during planning~\citep{janner2019trust,yu2020mopo}, we internalize this knowledge \emph{into the policy itself} as a cold-start initialization: by learning to describe the current state and predict its next state, the agent acquires a grounded prior over dynamics that it subsequently refines with RL. How to acquire such a prior effectively and scalably for LLM agents remains open. We answer this with \texttt{SPA} (Self-exPerience Agent)\camera{\footnote{\camera{Code, configuration files, and checkpoints: \url{https://github.com/shiqichen17/SPA}.}}}, a simple \textbf{explore-before-exploit} framework: the agent first gains self-experience to ground itself in the environment, then exploits this grounding to act effectively.

\texttt{SPA} focuses on exploration on both state space and action space: 
\textbf{1) State Estimation.} 
To explore the state space, we transform unfamiliar states to familiar ones. 
\textbf{2) Transition Modeling.} Exploration of the action space means making the environment's transitions predictable with an accurate transition kernel $p_\theta(s_{t+1}\mid s_t,a_t)$. 
\texttt{SPA} scales improvements in RL: both {Pass@1} and {Pass@k} grow over training. 
\begin{figure}[!t]
    \centering
    \includegraphics[width=0.85\columnwidth]{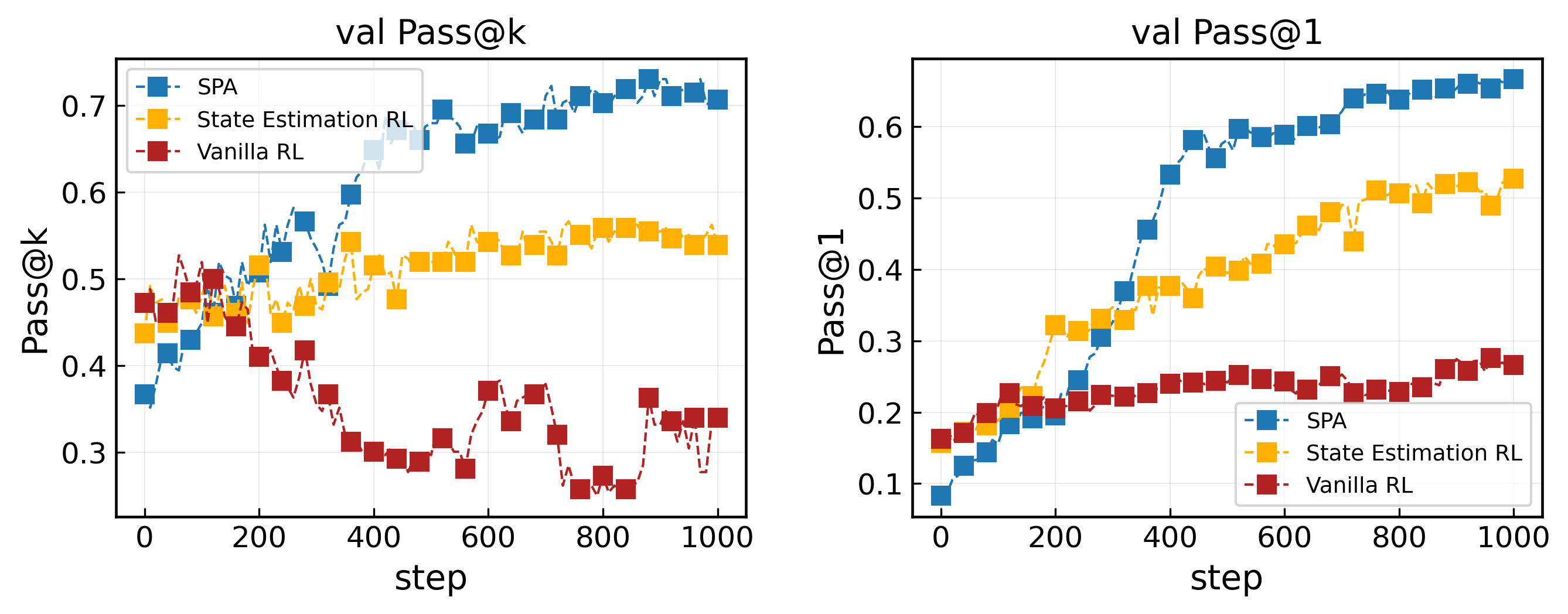}
    \vspace{-12pt}
    \caption{Validation performance on Sokoban and Frozen Lake. Red: Vanilla RL (StarPO baseline); Yellow: state-estimation RL; Blue: \texttt{SPA} (world-modeling SFT as a cold start, then state-estimation RL). Left: \emph{Pass@k}; Right: \emph{Pass@1}.}
    \label{fig:passk_curves}
    \vspace{-2pt}
\end{figure}
We evaluate \texttt{SPA} on two well-studied grid worlds, \emph{Sokoban} and \emph{FrozenLake}, and a math-reasoning environment, \emph{Sudoku}. These tasks demand multistep spatial reasoning and precise action control, which are unfamiliar for the pretrained model. Under a fixed budget of 1000 training steps, our method lifts average success rates on \emph{Sokoban} from 25.6\% to 59.8\% and on \emph{FrozenLake} from 22.1\% to 70.9\% (Figure~\ref{fig:passk_curves}). Qualitative case studies further show that the internal world model induces more deliberate, globally consistent plans that benefit subsequent RL. Compared against a reward-shaping world-modeling method under matched settings, \texttt{SPA} injects this grounding more effectively than adding a world-modeling reward during online RL~\citep{wang2025vagen}; we hypothesize that such an auxiliary reward interferes with the task objective rather than yielding reusable dynamics.

Finally, controlled analyses isolate which factors drive the gains. The improvements are not explained by additional finetuning alone, but track four ingredients: explicit transition modeling, grounded state representations, self-experience trajectories from a sufficiently strong exploration policy, and adequate coverage of transition data. Together they let the agent better interpret unfamiliar observations, model environment dynamics, and transfer to downstream policy optimization.

%% file: sections/3-collapse.tex
We find that agentic RL behaves very differently across environments. In some environments, reinforcement learning improves both single-trajectory success and broader success across multiple attempts. In contrast, in unseen environments such as Sokoban, FrozenLake, and Sudoku, performance degrades much more sharply. These environments present states that are much less familiar to the policy model, making them harder to parse and reason about. As shown in Table~\ref{tab:state-ppl}, Sokoban, FrozenLake, and Sudoku exhibit much higher state perplexity relative to the size of their underlying state space, whereas environments such as ALFWorld and WebShop are much easier for the model to process. This suggests that the difficulty is due to a mismatch between the environment states and the model's prior knowledge.
\camera{We compute this perplexity over the serialized state-representation tokens, so we interpret it as a proxy for the \emph{representational familiarity} of state strings to the LM, rather than as a calibrated measure of environment entropy or state-space size.}

\input{Tables/ppl.tex}
To better understand how RL fails in such environments, we examine two metrics throughout training: Pass@1, which measures the success rate of the highest-probability trajectory, and Pass@k, which measures the success rate across multiple sampled trajectories and thus reflects broader coverage over plausible solution paths. We observe a systematic divergence between these two metrics in unseen environments. As shown in Figure~\ref{fig:teaser}, in Sokoban, FrozenLake, and Sudoku, standard RL causes Pass@k to consistently deteriorate over training, while Pass@1 improves only marginally. This pattern is markedly different from training-aligned environments like ALFWorld and WebShop, where both metrics improve together, and pass@k fastly goes to 100 percent.

This divergence suggests that, in unseen environments, RL improves single-path exploitation without improving broader trajectory coverage. As training proceeds, the policy becomes narrower: it looks better at converging on one likely path, but fails to acquire the broader world knowledge needed for robust generalization. We therefore interpret the decline of Pass@k as evidence that exploration is collapsing. This observation motivates our paper: if unseen environments narrow exploration because the agent lacks grounded state and transition understanding, improving such grounding should help mitigate this collapse, which we probe in the next section.

%% file: Tables/ppl.tex
\begin{table}[t]
\centering
\small
\begin{tabular}{lrr}
\toprule
Task & PPL & \#States \\
\midrule
Sokoban & 163.9 & 7 \\ 
\textcolor{red}{~~+ State Estimation} & \textcolor{red}{19.6} & \\
Frozen Lake & 187.1 & 6  \\
\textcolor{red}{~~+ State Estimation} & \textcolor{red}{15.4} & \\
Sudoku & 15.5 &5  \\
\textcolor{red}{~~+ State Estimation} & \textcolor{red}{7.3} & \\
ALFWorld & 6.0 & $ |V| $ \\
WebShop & 11.7 & $ |V| $ \\
\bottomrule
\end{tabular}
\caption{Average PPL of state representations on Qwen2.5-Instruct-1.5B. 
\#States denotes the number of possible values each grid cell can take (=random guess PPL). }
\label{tab:state-ppl}
\end{table}

%% file: sections/3-method.tex
\begin{figure*}[t]
    \centering
    \includegraphics[width=0.95\textwidth]{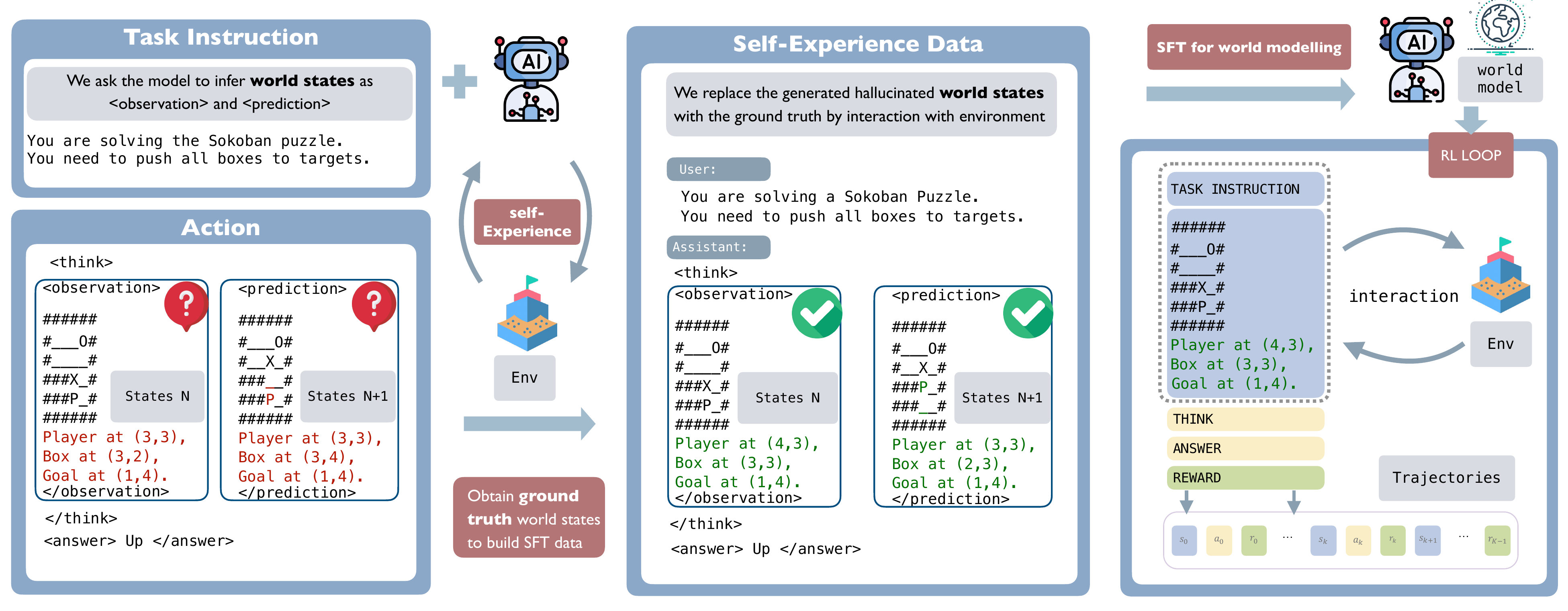} 
    \caption{~\texttt{SPA} equips an LLM agent with an inner world model. The world model is decomposed into two parts: (i) State Estimation, which converts raw observations into structured, informative natural-language descriptions of the world state; and (ii) Transition Modeling, trained in a self-experience Supervised Fine-Tuning setting to predict next-state representations from current states.}
    \label{fig:main}
\end{figure*}

\vspace{-10pt}
In this section, we introduce \texttt{SPA}, a simple and effective framework for improving LLM agent training. We show the framework in Figure~\ref{fig:main}. The central idea is to strengthen grounding through explicit world modeling prior to policy learning. Specifically, we decompose world modeling into two components: \textit{state representation} and \textit{transition dynamics}. By jointly modeling states and transition dynamics, the agent is regularized to follow structured state descriptions while interacting with the environment, thereby improving grounding. We train the world model via supervised finetuning on transition dynamics and use it to initialize policy learning. Our method does not rely on external knowledge or larger teacher models; instead, it learns entirely from the base model's own experience. In the following, we present the state-representation schema and the transition-modeling procedure. 

\subsection{Optimizing State Estimation}\label{sec:optimizing-state-repre}

We hypothesize that a key challenge in agent RL training is the distribution shift: observed states are unfamiliar relative to the model’s pretraining data, making them difficult to parse and reason about. This is especially pronounced when environment states are rendered as symbolic text strings. For example, a \(6\times6\) Sokoban board is serialized as \texttt{\#\#\#\#\#\#\symbol{92}n}\allowbreak\texttt{\#\#\#\#\#\#\symbol{92}n}\allowbreak\texttt{\#\_\_\#\#\#\symbol{92}n}\allowbreak\texttt{\#\_\_\#\#\#\#\symbol{92}n}\allowbreak\texttt{\#O\_XP\#\symbol{92}n}\allowbreak\texttt{\#\#\#\#\#\#} in~\citet{wang2025ragen}. As shown in Table~\ref{tab:state-ppl}, such raw representations have high perplexity, whereas environments with natural-language state descriptions, such as ALFWorld and WebShop, exhibit much lower perplexity.
\input{Tables/main_fig}
To reduce this mismatch, we augment the raw state \(s'_t\) with an abstract representation \(b_t\), giving \(s_t=\text{Concat}(s'_t, b_t)\). In text games, \(b_t\) gives coordinates of key entities (player, boxes, goals). This lowers perplexity (Table~\ref{tab:state-ppl}) and improves both Pass@k and Pass@1 over the \textit{Vanilla RL} baseline, as shown by \textit{State Estimation RL} in Table~\ref{tab:mainres}. We next introduce transition-dynamics modeling on this representation.


\subsection{Injecting Transition Modeling via Self-Experience Fine-tuning}
Based on the interpretable environment states, we inject transition dynamics into the LLM agent via a self-experience finetuning process. It includes two stages: ~\textit{trajectory collection via self-experience} and~\textit{dynamics transition training}. During the trajectory collection stage, we use the base model to freely explore and interact with the environment under a self-experience scenario to update the model's belief of the world. 

\paragraph{Self-Experience Data Generation} We collect data by prompting the model to interact with the environment. This interaction repeats over $T$ turns to collect a trajectory $\tau = (s_0, o_0, a_0, r_0, \ldots, s_{T}, o_{T})$. At each step, the action $a_t$ is parsed and executed. The environment then returns a reward $r_t$ and transitions to a new state $s_{t+1}$, which gives the agent a new observation $o_{t+1}$.
The structure of $a_{t}$ is designed to incorporate explicit visual state reasoning, with a particular focus on world modeling. To reason about world states, we require the model to think before taking actions: model generation $a_t=(
{z}_t, \hat{a}_t)$ is a sequence of text including both reasoning tokens $
z_{t}$ and executable action $\hat{a}_t$. 
To incorporate world modeling into the trajectories, we prompt the LLM agent to describe the current state and predict the next state during reasoning~\citep{wang2025vagen}. Specifically, the reasoning variable $z_{t}$ between  \tokenthinks ~and  \tokenthinke ~is required to explicitly represent both the current state and the predicted next state:
\begin{center}\small
\tokenthinks\allowbreak\tokencurs\allowbreak$\statebelief{t}$\allowbreak\tokencure\allowbreak\camera{[\textit{plan}]}\allowbreak\tokennexts\allowbreak$\statebelief{t+1}$\allowbreak\tokennexte\allowbreak\tokenthinke\allowbreak\tokenans\allowbreak$\hat{a}_t$\allowbreak\tokenane.
\end{center}
For the next step, we then replace the model's self beliefs about current states  $\statebelief{t}$ and future states $\statebelief{t+1}$ with the ground-truth states $s_t$ and $s_{t+1}$ from the environment, resulting in the training trajectory:
\begin{center}\small
\tokenthinks\allowbreak\tokencurs\allowbreak$\statetrue{t}$\allowbreak\tokencure\allowbreak\camera{[\textit{plan}]}\allowbreak\tokennexts\allowbreak$\statetrue{t+1}$\allowbreak\tokennexte\allowbreak\tokenthinke\allowbreak\tokenans\allowbreak$\hat{a}_t$\allowbreak\tokenane.
\end{center}
\camera{Here [\textit{plan}] denotes optional free-form planning text, and $\hat{a}_t$ may contain several primitive actions separated by ``\texttt{||}''. During SFT, we rewrite only the \tokencurs{} and \tokennexts{} spans with grounded states and keep plans and actions as generated; we list the exact per-environment prompts in Appendix~\ref{appdix: prompts}.}
Next, we perform supervised finetuning (SFT) on these trajectories that reflect the environmental dynamics, enabling the model to learn environment dynamics. The resulting SFT model serves as a strong initialization for downstream RL training.

\paragraph{World Modeling.} With the constructed trajectory $\tau_{1:T}$ above,
and let $\tau_{<i}$ denote the prefix up to token $i-1$, we optimize a simple masked token-level cross-entropy over tokens enclosed by the special tags to capture environmental dynamics and to internalize world models in agents.
\begin{align}
\label{eq:sft}
\mathcal{L}_{\text{W}}(\theta)
&= - \frac{1}{\sum_{i=1}^{T} M_{i}}
   \sum_{i=1}^{T} M_{i}
   \log p_{\theta}(\tau_i \mid \tau_{<i}), \\
\text{where}~ M_{i} &= \mathbbm{1}\big[\,\tau_i \in  \text{span}(\tokenthinks, \tokenthinke) \notag\\
&\quad \cup~ \text{span}(\tokenans, \tokenane)\,\big]. \notag
\end{align}
We explicitly represent both the prediction of the current state and the next state in the reasoning.  
This simple world-modeling objective improves the agent’s grounding by helping it generate better next states and answers. During training, tokens corresponding to environment observations are masked in the loss. The resulting model then provides an effective initialization for downstream policy RL.

\camera{We emphasize that \texttt{SPA} is a \emph{policy-centric} world-modeling objective, not a standalone action-conditioned simulator $p(s_{t+1}\mid s_t, a_t)$: the next-state prediction is emitted inside the same reasoning trace that produces the action, conditioned on a prefix (grounded current state, often an implicit plan) generated by the collecting policy, so the learned predictive distribution is tied to that policy. We therefore use the internalized world model as a grounded \emph{initialization} for RL rather than an inference-time simulator, and validate this in \S\ref{subsec:transition} and Appendix~\ref{appdx:analyses}.}

\subsection{RL Training}
After obtaining a good world model by SFT, we do RL to boost the policy model. We use the Proximal Policy Optimization (PPO)~\citep{schulman2017proximal} algorithm for RL. 
Denote $\tau$ as trajectory, $u_i = \frac{\pi_\theta(\tau_{i} |  \tau_{<i})}{\pi_{\text{old}}( \tau_{i}| \tau_{<i})}$ as the probability ratio between the current and old policies,
and let ${\tau}_{<i}$ denote the prefix of token $i$. The PPO loss is defined as:
\begin{equation*}
\begin{aligned}
J^{\text{PPO}}(\theta) ={}& \frac{1}{\sum_i M_i} \sum_i M_i \cdot \min\big( u_i A_i, \\
& \quad \text{clip}(u_i, 1 - \varepsilon, 1 + \varepsilon)\, A_i \big),
\end{aligned}
\end{equation*}
where $M_i^{}$ is a mask that is $1$ for answer tokens and $0$ for contextual tokens, $A_i$ is the per-token advantage and $\varepsilon$ is a clipping hyperparameter. This objective is to optimize the policy to gain more rewards. Here we use our optimized $s_t$ (described in Section~\ref{sec:optimizing-state-repre}) to represent the world states to better encode the world knowledge during RL training.

%% file: Tables/main_fig.tex
\begin{table*}[t]\centering
\small
\begingroup
\setlength{\tabcolsep}{4pt}
\renewcommand{\arraystretch}{0.88}
\scalebox{0.92}{
\begin{tabular}{>{\raggedright\arraybackslash}p{7.2cm}cccccc}
\toprule
\multirow{2}{*}{\textbf{Model}} & \multicolumn{2}{c}{\textbf{Sokoban}} & \multicolumn{2}{c}{\textbf{FrozenLake}} & \multicolumn{2}{c}{\textbf{Sudoku}} \\
\cmidrule(lr){2-7}
& \textbf{Pass@1} & \textbf{Pass@8} & \textbf{Pass@1} & \textbf{Pass@8} & \textbf{Pass@1} & \textbf{Pass@8} \\
\midrule
Qwen2.5-0.5B-Instruct         & 5.5  & 25.8 & 8.1  & 32.8 & 0.0  & 0.0  \\
+Vanilla RL                   & 16.9 & 35.9 & 22.8 & 30.1 & 0.0  & 0.0  \\

+State Estimation RL         & 20.4 & 38.7 & 24.2 & 26.6 & 4.5  & 14.1 \\
+World Model RL (VAGEN adaptation)                         & 33.3 & 44.9 & 25.8  & 40.6  & 0.1  & 0.8  \\
+~\texttt{SPA}                      &
\cellcolor{MyCellShade}36.7 & \cellcolor{MyCellShade}45.3 &
\cellcolor{MyCellShade}46.9 & \cellcolor{MyCellShade}65.6 &
\cellcolor{MyCellShade}18.2 & \cellcolor{MyCellShade}43.8 \\

\midrule
Qwen2.5-1.5B-Instruct         & 16.3 & 47.3 & 17.2 & 49.2 & 1.6  & 11.3 \\
+Vanilla RL                   & 25.6 & 34.0 & 22.1 & 30.7 & 0.0  &  0.0 \\

+State Estimation RL         & 52.7 & 53.9 & 27.6 & 34.8 & 39.1 & 72.3 \\
+World Model RL (VAGEN adaptation)                      &  44.5  &  50.0  &  37.7   &  43.0  &  0.0   &  0.0   \\
+~\texttt{SPA}                      &
\cellcolor{MyCellShade}59.8 & \cellcolor{MyCellShade}69.5 &
\cellcolor{MyCellShade}70.9 & \cellcolor{MyCellShade}75.0 &
\cellcolor{MyCellShade}59.6 & \cellcolor{MyCellShade}94.9 \\

\midrule
Qwen2.5-3B                    & 12.5 & 35.9 &  6.9 & 26.9 & 0.0  &  0.0 \\
+Vanilla RL                   & 31.4 & 35.5 &  7.6 & 16.0 & 0.0  &  0.0 \\
+State Estimation RL         & 26.2 & 27.7 & 24.7 & 35.5 & 24.1 & 26.2 \\
+World Model RL (VAGEN adaptation)                        &  31.9  &  43.0   &  30.7   &  34.0   &  0.0  &  0.0   \\
+~\texttt{SPA}                      &
\cellcolor{MyCellShade}49.7 & \cellcolor{MyCellShade}58.2 &
\cellcolor{MyCellShade}41.3 & \cellcolor{MyCellShade}46.1 &
\cellcolor{MyCellShade}69.9 & \cellcolor{MyCellShade}89.8 \\

\midrule
LLaMA3.2-1B-Instruct          &  8.3 & 33.2 & 10.2 & 41.0 &  0.0 &  0.0 \\
+Vanilla RL                   & 21.2 & 39.8 & 10.8 & 24.6 &  0.1 &  1.2 \\
+State Estimation RL         & 31.6 & 44.1 & 19.3 & 29.7 &  45.0&  71.1\\
+World Model RL (VAGEN adaptation)                        & 47.4 & 50.8 & 24.5 & 29.3 &  0.1 & 0.8  \\
+~\texttt{SPA}                      &
\cellcolor{MyCellShade}53.0 & \cellcolor{MyCellShade}68.0 &
\cellcolor{MyCellShade}64.8 & \cellcolor{MyCellShade}71.1 &
\cellcolor{MyCellShade}81.3 & \cellcolor{MyCellShade} 100 \\

\midrule
GPT-OSS-20B                   & 45.8 & 84.4 & 68.8 & 100  &  61.8& 100  \\
+State Estimation            & 55.1 & 89.1 & 73.3 & 100  &  68.1& 100  \\
\bottomrule
\end{tabular}

}

\endgroup
\caption{Results on Sokoban/FrozenLake/Sudoku for different models. (Metrics in $\times 10^{-2}$).}
\label{tab:mainres}
\end{table*}

%% file: sections/4-exp.tex
\vspace{-3pt}
\paragraph{Experiment setting and baselines}We conduct experiments on a diverse set of models of varying sizes, including Qwen2.5-0.5B/1.5B-Instruct, Qwen2.5-3B~\citep{qwen2.5} and LLaMA-3.2-1B-Instruct~\citep{dubey2024llama}. Our evaluation spans multiple environments like Sokoban, FrozenLake, and Sudoku, all of which are highly controllable. Our codebase is developed based on RAGEN~\citep{wang2025ragen}. 
We train with a batch size of 32 and collect eight rollouts per environment at each update step. For evaluation, we use both Pass@8 and Pass@1 metrics. We use VAGEN~\citep{wang2025vagen} as our baseline. This method enables online world-model learning during RL by providing observation and prediction-based reward signals.  We also evaluate a SOTA model: open-source GPT-OSS-20B~\citep{openai2025gptoss120bgptoss20bmodel}.
\paragraph{Small models could outperform strong models on text games with \texttt{SPA}.}
Our main results are shown in Table~\ref{tab:mainres}. For every model we evaluated, \texttt{SPA} consistently outperforms State Estimation RL, which in turn outperforms vanilla RL. E.g, for Qwen2.5-1.5B-Instruct,~\texttt{SPA} could obtain 59.8 Pass@1 score after 1000 training steps for Sokoban environment, and 70.9 Pass@1 for FrozenLake environment, which even outperforms GPT-OSS-20B model. Across other models, Qwen2.5-0.5B-Instruct and Qwen2.5-3B,~\texttt{SPA} delivers substantial improvements over baselines on all our evaluated environments. This is consistent with \texttt{SPA} restoring exploration---reflected in the recovered Pass@k---and consolidating environment knowledge in these settings. To probe whether \texttt{SPA} extends to more realistic settings, we also evaluate it on ALFWorld using the Qwen2.5-0.5B-Instruct model.
We choose a smaller 0.5B model as we observe that the performance quickly saturates for the 1.5B model (in Figure~\ref{fig:teaser}). 
The results are in Table~\ref{tab:0.5b-alfworld-passk-results}. 
We observe that \texttt{SPA} delivers consistent gains on both the Pass@1 and Pass@k metrics.
This offers initial evidence that \texttt{SPA} extends beyond text games, though here we verify it on a single model and leave broader real-world validation to future work.
\vspace{-3pt}
\paragraph{\texttt{SPA} outperforms the reward-based method VAGEN.}
VAGEN~\citep{wang2025vagen} performs world modeling by eliciting current-state observations and predicting next states, and then applies a reward to encourage accurate grounding. For a fair comparison, we reimplemented VAGEN in the same state-estimation setting as \texttt{SPA}\camera{, following the original formulation closely (prompt format and observation/prediction reward supervision)}. In our reimplementation under these matched settings, VAGEN yields little improvement on our testbed, whereas \texttt{SPA} delivers substantial gains \camera{(e.g., 59.8/69.5 vs.\ 44.5/50.0 on Sokoban, Table~\ref{tab:mainres})}. We attribute this, tentatively, to \texttt{SPA} confining world-model learning to the SFT stage and leaving the RL objective unperturbed, thereby avoiding the multi-objective interference of adding a world-modeling reward during RL.


%% file: sections/5-ablation.tex
We further disentangle why \texttt{SPA} succeeds, highlighting four critical factors: (i) \emph{transition modeling}, (ii) \emph{grounded state representations}, (iii) \emph{a strong policy for world-model interaction}, and (iv) \emph{sufficient world-modeling data}.

\subsection{Effect of Transition Modeling}
\label{subsec:transition}
\begin{figure}[t]
    \centering
    \includegraphics[width=1\columnwidth]{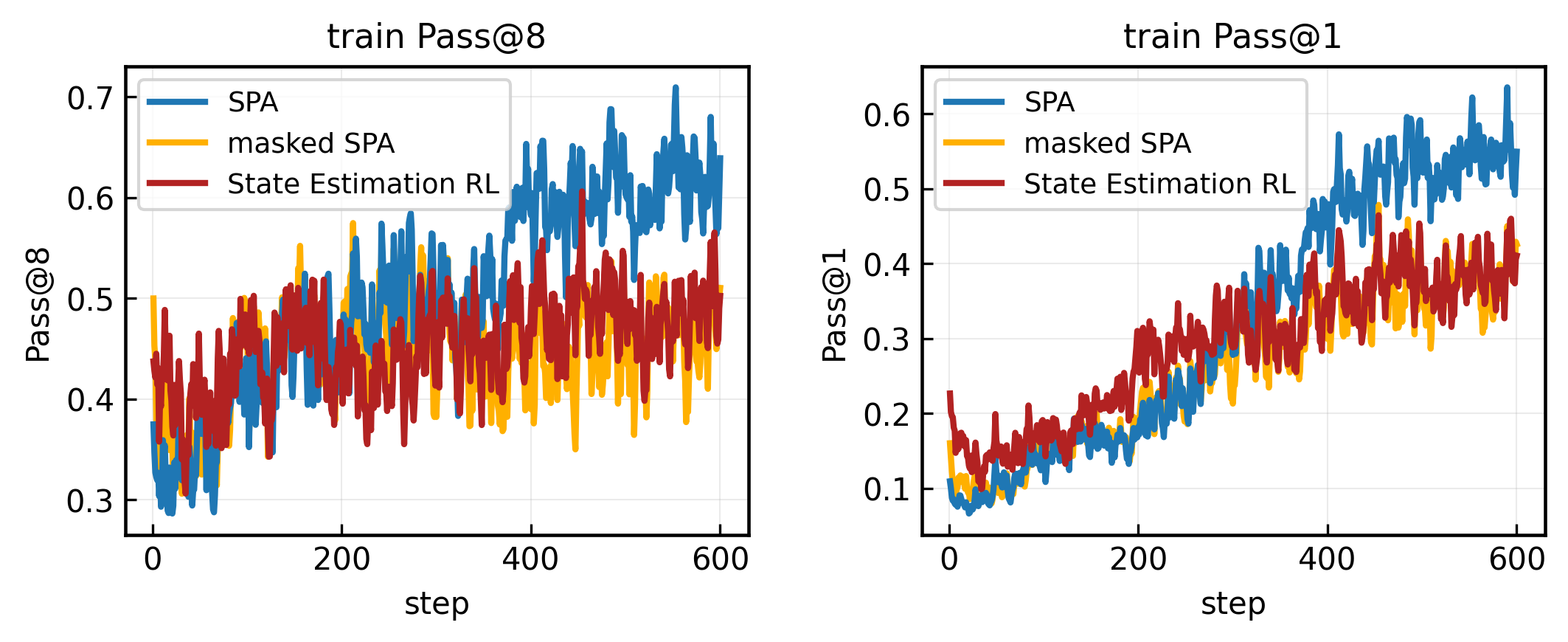} 
    \caption{RL performance after SFT under three settings: State Estimation RL, No masking vs. Masking current states with next states for SPA.}
    \label{fig:mask}
\end{figure}
\paragraph{Transition-model learning is key to RL scaling. Masking the SFT loss on the current and next states results in no improvement.} To isolate the effect of world modeling supervision, we remove explicit transition (world modeling) annotations (\textit{observation} and \textit{prediction} spans) from the SFT corpus and replace them with a masking token ~\textit{[MASKED]} and set the masking tokens' loss to 0 during SFT. Starting from the base pretrained model, we run SFT on this masked data and use the resulting checkpoint to initialize the PPO stage. We show the results in Figure~\ref{fig:mask}. In the masked setting, SFT yields no improvement for PPO. This suggests that PPO’s scaling benefits derive from transition-model learning, highlighting our world-model encoding stage as pivotal to the performance.\input{Tables/noreplace}

\paragraph{What happens if SFT relies solely on self-belief states rather than ground-truth states?}
To test whether ground-truth states are necessary, we train SFT using only the model’s self-belief states (i.e., without ground-truth supervision). This variant fails to improve, and in fact, degrades RL performance (Table~\ref{tab:noreplace}), indicating that accurate world modeling during SFT, not merely adding a state-prediction signal, is necessary for downstream gains.

\input{Tables/wo_abstraction}

\paragraph{Without State Estimation, the world modeling SFT could also boost the performance.} To verify whether the world-modeling SFT also boosts RL performance without State Estimation, we use a similar observation-then-prediction prompt to generate self-experience data from raw symbolic states. We observe a performance gain, as shown in Table~\ref{tab:woabs}. Thus, ~\texttt{SPA} improves RL even without explicit state estimation, further validating its effect.

\camera{We present two further confound-controlled analyses in Appendix~\ref{appdx:analyses}. First, dropping the next-state prediction while keeping the free-form plan lowers Pass@1/Pass@8 from 36.3/40.6 to 25.6/34.0, so the prediction slot is not redundant with planning. Second, on-policy prediction accuracy rises from 31.9\% to 46.6\% during RL, so the policy refines rather than ignores its predictions.}
\subsection{Effect of Ground Truth state abstraction}

\begin{figure}[t]
    \centering
    \includegraphics[width=0.9\columnwidth]{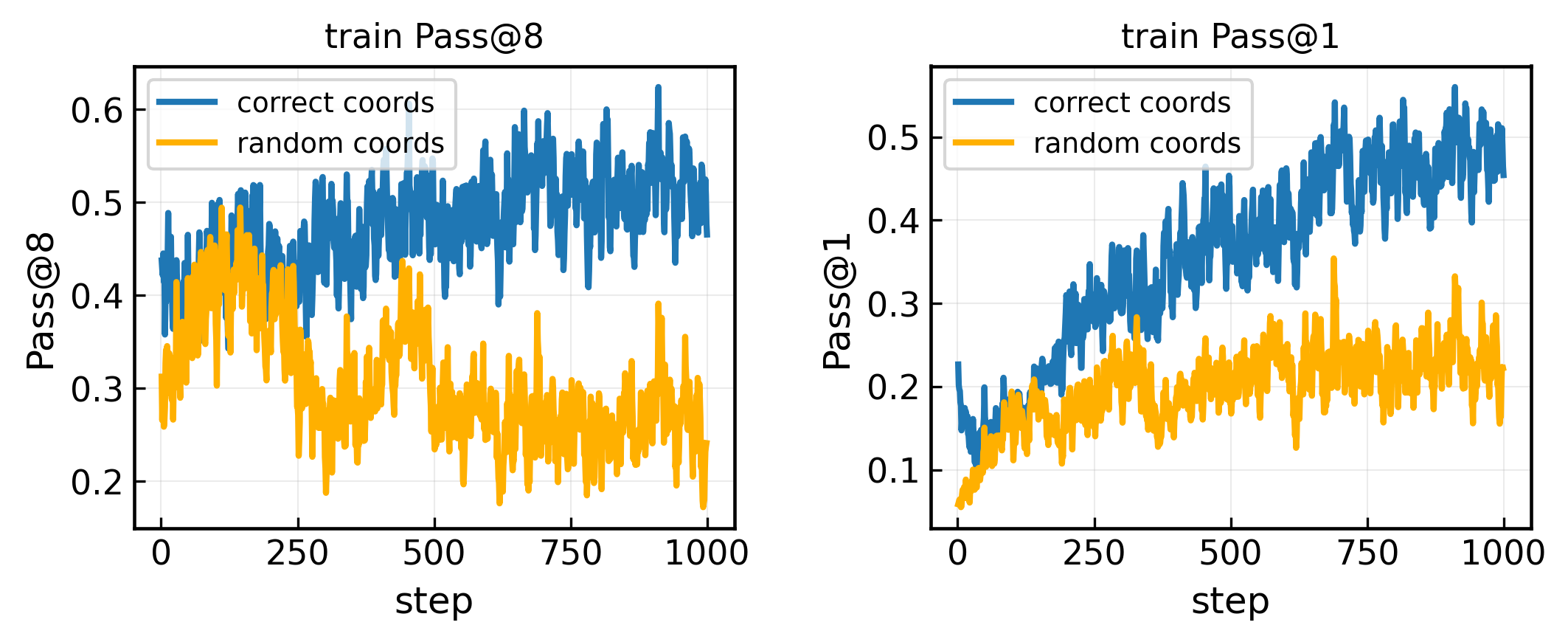} 
    \caption{Pass@8/Pass@1 comparison by using correct and incorrect abstraction on Sokoban during RL training. Left: Abstraction PPO with correct coordinates. Right: randomized coordinates.}
    \label{fig:random}
\end{figure}

\paragraph{Ground-truth coordinates are critical for RL training.} To test whether the abstract state format alone is sufficient, we hold the text template and replace all ground-truth $(x,y)$ coordinates (player/box/goal) with i.i.d. random coordinates drawn from the same grid at each step. This preserves the correct format but modifies the content. In Figure~\ref{fig:random}, with random coordinates, the training collapses: Pass@1 and Pass@8 stagnate at low levels with fluctuation, whereas with correct coordinates, both metrics improve steadily. It shows that coordinates provide essential supervision for spatial grounding and credit assignment; without them, the policy fails to align with actionable spatial relations, resulting in brittle exploration and poor generalization.

\subsection{Effect of Policy Prior for Exploration}
\paragraph{Can the RL policy be replaced with random actions for generating world-modeling trajectories?}
\input{Tables/random_action}
We realize that world-modeling trajectories need not be collected exclusively with the learned policy; they can also be obtained via generic action generators. To ablate \texttt{SPA}, we replace its SFT data generator with a random‑action generator: at each step, we sample uniformly from the action space ${\texttt{Up},\texttt{Down},\texttt{Left},\texttt{Right}}$ and substitute a fixed reasoning token sequence (\emph{“I will push the box to the target.”}) As shown in Table~\ref{tab:random-action}, trajectories produced by this random policy yield substantially worse downstream RL performance, suggesting that a prior-aware policy for self-experience data collection is crucial for effective world-model training to empower the subsequent RL.
\vspace{-3pt}
\paragraph{How to enable effective exploration when the policy model is weak?}
We find that many failures are due to instruction noncompliance during data generation: some models fail to follow the specified output format, preventing reliable synthesis of world-modeling samples. To quantify this effect, we conduct an ablation comparing performance before and after filtering out SFT samples that deviate from the required format. Building on this, we evaluate three settings under the same RL budget on the LLaMA-3.2-1B-Instruct model in Sokoban: \textbf{Original SFT}, distills world-state transitions via SFT; \textbf{Filtered SFT}, where the data samples are cleaned by enforcing minimal structural validity (\texttt{<observation>}–\texttt{<prediction>} reasoning followed by an \texttt{<answer>}); and \textbf{Abs PPO}, where PPO is trained on raw grids enriched by abstraction, without transition-model SFT. As in Figure~\ref{fig:filter-bar}, removing misaligned data improves training stability and performance (e.g., higher {Pass@k} and {@1}), indicating that format adherence is critical for learning accurate state and transition structure. 
\subsection{Effect of Extended SFT}
\begin{figure*}[!t]
    \centering
    \includegraphics[width=\textwidth]{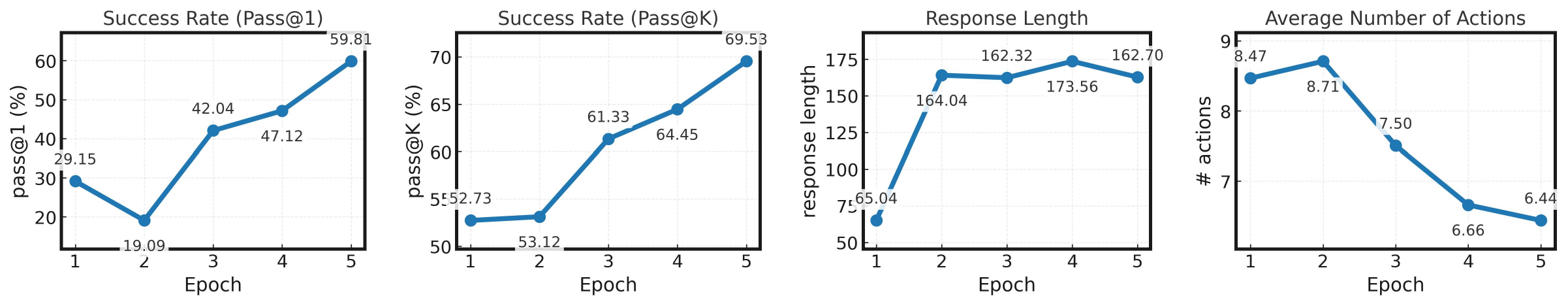} 
    \caption{We evaluate RL training under initializations with varying SFT epochs and observe that increasing world-modeling SFT generally improves subsequent RL performance. }
    \label{fig:multiepochs}
\end{figure*}

To investigate the impact of spending more compute on world-modeling SFT, we vary the length of SFT from 1 to 5 epochs, and perform RL on all 5 checkpoints under the same setup. 
Figure~\ref{fig:multiepochs} shows the final performance after 1000 steps of RL training. 
As the world model is trained longer, downstream RL improves consistently: Pass@1 rises from {0.29} to {0.60} and Pass@k from 0.53 to 0.70. Efficiency also improves: the average number of actions per episode decreases from 8.47 to 6.44, while action effectiveness rebounds from an early dip at Epoch~2's 0.60 to 0.77 by Epoch~5. Response length climbs from 65 tokens at Epoch~1 and then stabilizes in the 160 to 175 range for Epochs~2--5, indicating that gains are not caused by verbosity but by better spatial grounding and transition prediction.

\vspace{-8pt}

%% file: Tables/noreplace.tex
\begin{table}[t]
\centering
\small
\begin{tabular}{lll}
\toprule
Method & Pass@1 & Pass@8\\
\midrule
Vanilla RL & 25.6  & 34.0 \\
State Estimation RL & 52.7   & 53.9 \\
SPA  & 59.8 & 69.5 \\
SPA (w/ self-belief SFT) & 39.2 & 44.2 \\
\bottomrule
\end{tabular}
\caption{We replace the ground-truth states in the self-experience SFT data with the model's own self-belief states (i.e., without ground-truth supervision) and compare it with SPA. Relying on self-belief states degrades downstream RL performance. }
\label{tab:noreplace}
\end{table}

%% file: Tables/wo_abstraction.tex
\begin{table}[t]
\centering
\small
\begin{tabular}{lll}
\toprule
Method & Pass@1 & Pass@8\\
\midrule
Vanilla RL & 25.6  & 34.0 \\
SPA w/o S Estimation & 36.3   & 40.6 \\
\bottomrule
\end{tabular}
\caption{\texttt{SPA} without the State Estimation and compare with the baselines. }
\label{tab:woabs}
\end{table}

%% file: Tables/random_action.tex
\begin{table}[t]
\centering
\small
\begin{tabular}{lll}
\toprule
Method & Pass@1 & Pass@8\\
\midrule
Vanilla RL & 25.6  & 34.0 \\
State Estimation RL & 52.7   & 53.9 \\
SPA (1 epoch SFT) & 29.2~\downbad{23.5} & 52.7~\downbad{1.2} \\
SPA (5 epoch SFT) & 59.8~\up{7.1} & 69.5~\up{16.6} \\
SPA (1 epoch RandSFT) & 0.1~\downbad{52.6} & 0.4~\downbad{53.5}\\
SPA (5 epoch RandSFT) & 20.2~\downbad{32.5} & 50.0~\downbad{3.9} \\
\bottomrule
\end{tabular}
\caption{We replace the self-experience SFT data generator with a random-action generator and compare its performance with \texttt{SPA}. }\vspace{-2mm}
\label{tab:random-action}
\end{table}

%% file: sections/6-findings.tex
\section{Findings: Exploration, Exploitation, and Generalization}
\label{sec:findings}
We now examine training dynamics and generalization. Multi-turn RL reveals a distinctive exploration-exploitation trajectory, and easy-to-hard \texttt{SPA} generalizes to higher task complexities within the same environment family, though it fails to transfer across games with fundamentally different dynamics.
\begin{figure}[t!]
  \centering
  \includegraphics[width=\columnwidth]{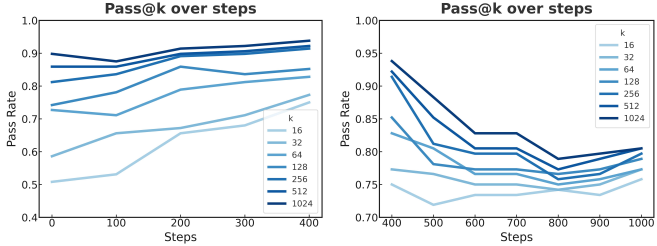}
  \caption{Pass@k performance on Sokoban during RL training. Left: steps 0--400, Pass@k consistently rises (exploration). Right: steps 400--1000, Pass@k declines as $k$ increases (exploitation). Even at $k=1024$, early exploration improves success rates.}
  \label{fig:passk-large}
\end{figure}

\begin{figure}[t!]
  \centering
  \includegraphics[width=0.8\columnwidth]{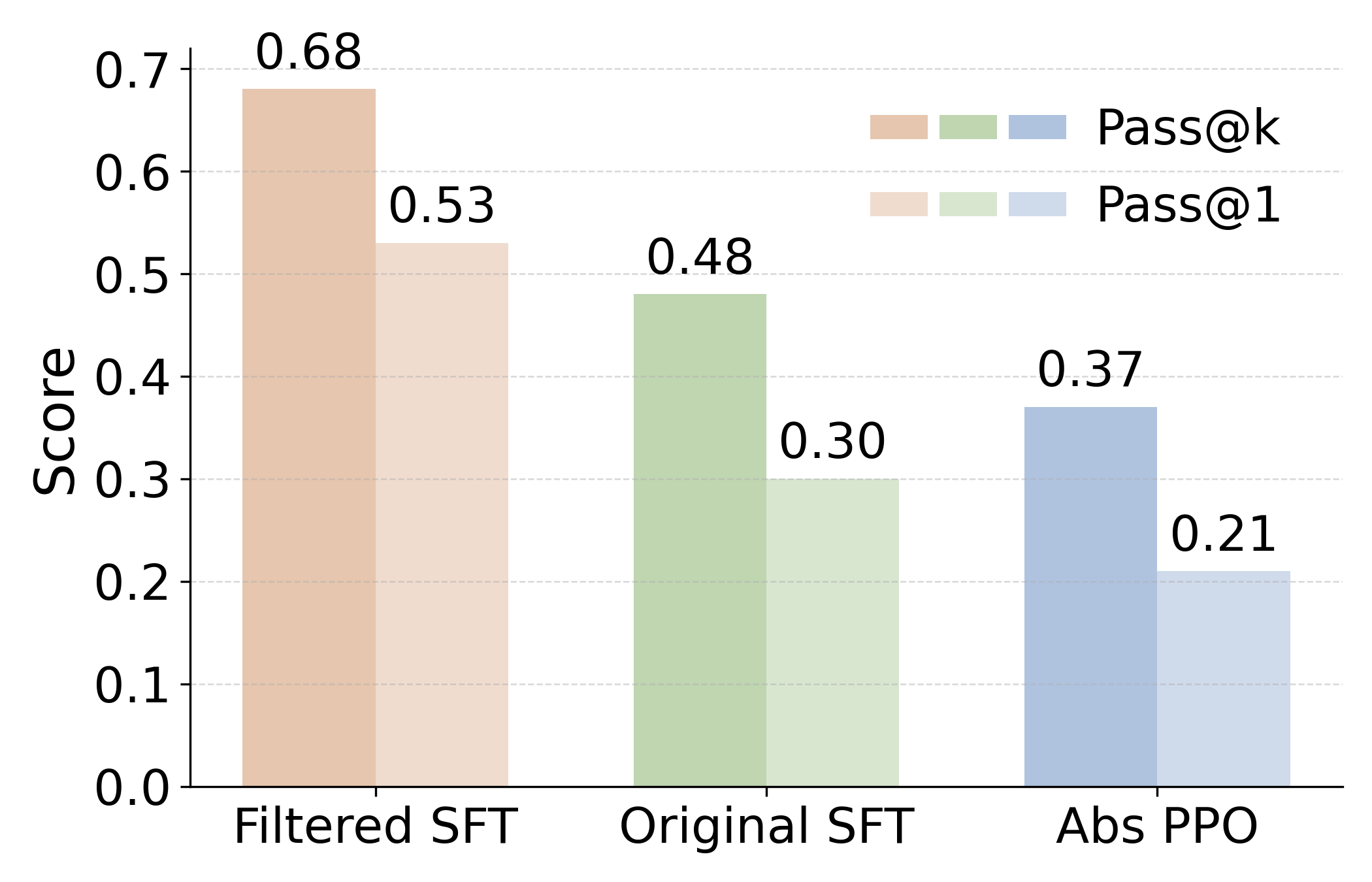}
  \caption{Results on {Sokoban} with LLaMA under three settings: Filtered~\texttt{SPA} SFT, raw data~\texttt{SPA} SFT and State Estimation PPO.}
  \label{fig:filter-bar}
  \vspace{-10pt}
\end{figure}
\paragraph{Environment interaction expands reasoning frontiers.}
By enabling direct environment interaction, multi-turn RL broadens the reasoning frontier. Prior work \citep{yue2025does} shows that RL with Verifiable Rewards (RLVR) improves sampling efficiency toward correct solutions but does not yield fundamentally new reasoning strategies. Moreover, RLVR exhibits a crossover in performance: RLVR-trained models outperform their base counterparts at small $k$ (e.g., $k=1$), while base models achieve higher Pass@k at large $k$. In contrast, our multi-turn setting displays no such reversal. As shown in Figure~\ref{fig:passk-large}, Pass@k continues to increase in the early phase of training (steps 0--400), remaining high even at $k=1024$, and only decreases during the later phase (steps 400--1000). This trajectory reflects an exploration-then-exploitation dynamic of the policy, highlighting the role of environment exploration and offering a plausible explanation for the discrepancy from math-domain RLVR results.

\camera{\paragraph{Beyond Pass@k.} We complement Pass@k with direct trajectory metrics in Appendix~\ref{appdx:analyses}: we find that grounded \texttt{SPA} runs visit \emph{fewer} total states than vanilla runs, yet cover far more success-relevant states with higher action diversity. Our claim is therefore not ``more exploration'' in a naive coverage sense, but \emph{more useful exploration}.}

\paragraph{World-modeling enables easy-to-hard transfer gains.}
We investigate whether world modeling learned in a low-complexity setting can generalize to a more complex setting and thereby accelerate RL. Concretely, we train the world model with SFT on the simple FrozenLake environment (grid size \(4\times4\)) and then perform RL on the more challenging variant (\(6\times6\)). In Figure~\ref{fig:easy2hard}, the proposed easy-to-hard transfer substantially outperforms all non-world-modeling baselines, yielding faster learning and higher asymptotic returns. Thus, learning simple world dynamics serves as a reusable prior that generalizes to more complex settings.
\camera{We note that this is a \emph{transfer analysis} rather than a same-budget comparison: easy-to-hard \texttt{SPA} receives extra pre-RL signal (SFT data from the easy variant) that the baselines do not use.}


\paragraph{Generalization across games and complexity levels.}
We further examine whether the trained game agents could generalize beyond its training setting. Table~\ref{tab:cross} reports two evaluation protocols. 
In the \emph{cross-complexity} setting, \texttt{SPA} is trained on a simple Sokoban variant ($6{\times}6$ grid, one box) and tested on a more complex variant ($10{\times}10$ grid, two boxes). \texttt{SPA} outperforms a baseline trained directly on the complex task, indicating that world-model SFT facilitates scaling to higher task complexity within the same environment family.
In the \emph{cross-game} setting, \texttt{SPA} trained on Sokoban is evaluated on FrozenLake environment, indicating that cross-game generalization is difficult.  

\input{Tables/crossgame}

%% file: Tables/crossgame.tex
\begin{table}[t]
  \centering
  \setlength{\tabcolsep}{6pt}
\vspace{-5pt}
  \begin{adjustbox}{max width=\columnwidth}
    \begin{tabular}{@{}lcc@{}}
      \toprule
      \textbf{Evaluation setting} & \textbf{Pass@1 (\%)} & \textbf{Pass@k (\%)} \\
      \midrule
      Simple $\rightarrow$ Complex (Sokoban) & 0.9  & 3.1  \\
      Complex (Sokoban, baseline)            & 0.1  & 0.8  \\
      \midrule
      Sokoban $\rightarrow$ FrozenLake       & 15.9 & 49.2 \\
      FrozenLake (baseline)                  & 17.2 & 49.2 \\
      \bottomrule
    \end{tabular}
  \end{adjustbox}

  \caption{Results of cross-complexity and cross-game generalization experiments. The upper two rows are cross-complexity, and the lower two rows are cross-game generalization. We see that trained agents can generalize to harder environments but not to other environments.}
  \label{tab:cross}
\end{table}

\begin{figure}[t]
  \centering
  \vspace{-5pt}
  \includegraphics[width=0.8\columnwidth]{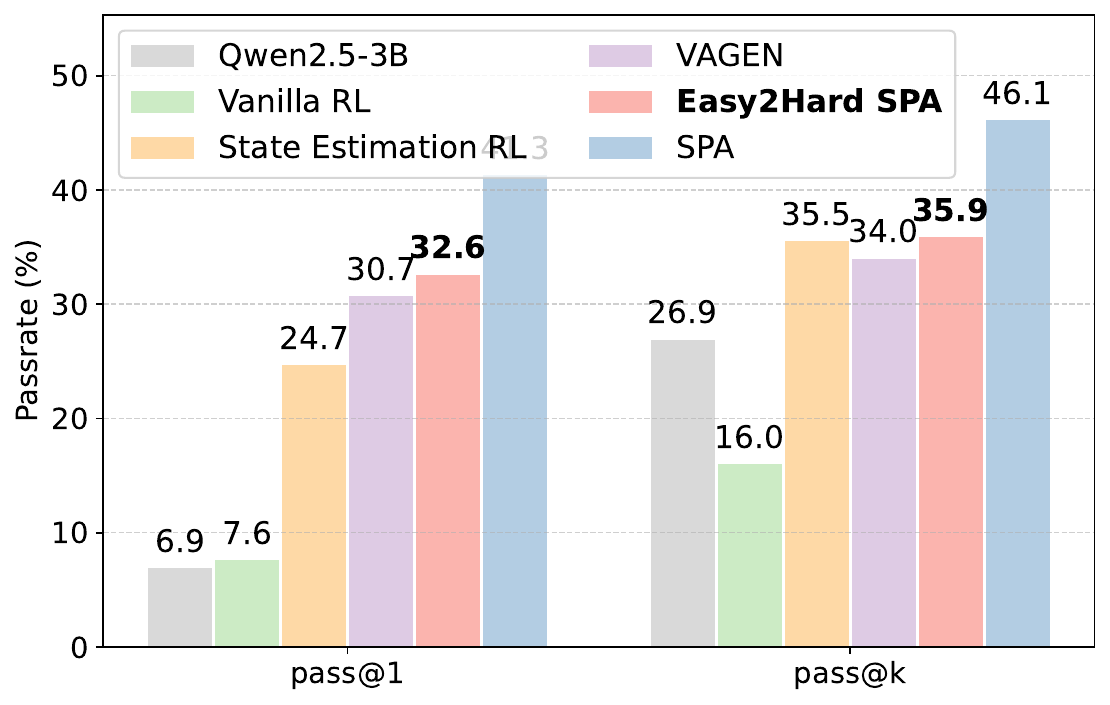}
  \caption{Easy-to-hard SPA on FrozenLake surpasses all baselines, trailing SPA trained on the target complexity.}
  \label{fig:easy2hard}
  \vspace{-5pt}
\end{figure}


%% file: sections/7-relwork.tex
\begin{camerablock}
\textbf{World modeling for LLM agents.} LLM agents trained via prompting~\citep{yao2023react,shinn2023reflexion} or RL pipelines~\citep{song2024trial,wang2025ragen} follow a short-horizon ``think-then-act'' paradigm with no internal mechanism to persist or simulate environment dynamics, and degrade outside the pre-training distribution~\citep{shao2019survey,szot2023large,ning2025survey,xing2025critiquesworldmodels}. Classical RL likewise shows the benefits of explicit world models~\citep{janner2019trust,yu2020mopo}; we internalize state and transition modeling into the policy as a grounded initialization. Closest to us, VAGEN~\citep{wang2025vagen} injects world-modeling signals into online RL via reward shaping, whereas \texttt{SPA} confines world modeling to an SFT stage, leaving RL unperturbed.

\textbf{Exploration and Pass@k dynamics in RL-trained LLMs.} The Pass@1/Pass@k divergence has been discussed for single-turn RLVR: \citet{yue2025does} show that RLVR improves small-$k$ success while narrowing the large-$k$ boundary, \citet{wen2025reinforcement} argue Pass@k can be incomplete, and \citet{zhai2026does} analyze tool-use agents with a Pass@$(k,T)$ metric. We do not claim the divergence is new; rather, we characterize its \emph{unfamiliar multi-turn} manifestation, give a grounding diagnosis with direct exploration measurements (\S\ref{sec:findings}), and mitigate it with a grounded initialization.
\end{camerablock}


%% file: sections/8-conclusion.tex
\section{Conclusion}
We identify \emph{exploration collapse}---a decline in Pass@k under RL in environments whose states are unfamiliar to the policy---and trace it to weak grounding in environment states and dynamics. \texttt{SPA} addresses this with a simple explore-before-exploit recipe: a self-experience SFT stage that grounds the agent in state estimation and transition prediction, used to initialize standard RL. It consistently outperforms vanilla RL and a reward-shaping baseline across four model families, indicating that grounding the policy before reward optimization---not additional finetuning per se---restores exploration. We hope this phenomenon and simple remedy offer a reproducible starting point for agents in unfamiliar environments.


%% file: sections/appendix-prompts.tex

\begin{promptbox}{Sokoban Prompt: Symbol with Coordinates}
You are solving the Sokoban puzzle. 
You are the player and you need to push all boxes to targets. You are provided with a symbol grid and (*@\textcolor{red}{the zero-indexed coordinates of the player, each box, and each target. Coordinates range from the top-left corner (0, 0) to the bottom-right corner (5, 5).}@*) 
...
The meaning of each symbol in the state is: 
#: wall, _: empty, O: target, √: box on target, X: box, P: player, S: player on target.
Your available actions are: Up, Down, Left, Right
...
Turn 1:
State:
###### 
#___O# 
#____# 
###X_# 
###P_# 
######
(*@\textcolor{red}{Player (P) is at (4,3); box (X) is at (3,3); target (O) is at (1,4).}@*)
...
\end{promptbox}

\begin{promptbox}{Sokoban Prompt: Base mode}
You are solving the Sokoban puzzle.
You are the player and you need to push all boxes to targets. 
You are provided with a symbol grid and the zero-indexed coordinates of the player, each box, and each target. 
Coordinates range from the top-left corner (0, 0) to the bottom-right corner (5, 5). 
When you are exactly next to a box, you can push it by moving in the same direction. 
You cannot push a box through a wall, and you cannot pull a box. 
The answer should be a sequence of actions, like <answer>Right || Right || Up</answer>.
\end{promptbox}



\begin{promptbox}{Sokoban Prompt: Observation then prediction}
You are solving the Sokoban puzzle.
You are the player and you need to push all boxes to targets. 
You are provided with a symbol grid and the zero-indexed coordinates of the player, each box, and each target. 
Coordinates range from the top-left corner (0, 0) to the bottom-right corner (5, 5). 
When you are exactly next to a box, you can push it by moving in the same direction. 
You cannot push a box through a wall, and you cannot pull a box. 
The answer should be a sequence of actions, like <answer>Right || Right || Up</answer>.

A sample full output is as follows:  
<think> 
<observation>
######
#_####
#_P###
#_X#_#
#__O_#
######  
Player (P) is at (2,2); box (X) is at (3,2); target (O) is at (4,3).  
</observation> 
1 Down - I push box to (4,2). 
2 Left - I step to (3,1).  
3 Down - I stand left of box, ready to push it Right onto target.  
<prediction>
######
#_####
#__###
#__#_#
#PXO_#
######
</prediction> 
</think>  
<answer> Down || Left || Down </answer>
\end{promptbox}

\begin{promptbox}{FrozenLake Prompt: Base mode}
You are solving the FrozenLake puzzle. 
Forbid the whole and go to the target. 
You may move to the unintended direction due to the slippery ice. 
Example answer format: 
<think>To forbid the hole and go to the target, I should go left then go up.
</think>
<answer>Left || Up</answer>
\end{promptbox}

\begin{promptbox}{FrozenLake Prompt: Observation then prediction}
You are solving the FrozenLake puzzle. 
Forbid the whole and go to the target. 
You may move to the unintended direction due to the slippery ice. 
Example answer format: 
<think>To forbid the hole and go to the target, I should go left then go up.
</think>
<answer>Left || Up</answer>

A sample full output is as follows: 
<think>
<observation>
_O__
O___
G___
__P_
</observation> 
Player at (3,2); holes at (0,1) and (1,0); goal at (2,0). 1 Up - move to safe ice (2,2). 2 Left - slide to (2,1), adjacent to goal. 3 Left - reach goal (2,0); player now on G.
<prediction>
_O__
O___
√___
____
</prediction>
</think> 
<answer> Up || Left || Left </answer>
\end{promptbox}

\begin{promptbox}{Sudoku Prompt: Base mode}
You are solving 4x4 Sudoku. 
Fill empty cells with digits 1-4.
Use a 1-indexed grid (rows/cols start at 1). 
A move is exactly: row,col,value (three integers). 
In one turn you may output multiple moves, separated by ||. 
Only propose moves that keep the row, column, and 2x2 subgrid valid. Always output EXACTLY as: 
<think>[brief reasoning]</think> 
<answer>[r,c,v || r,c,v ...]</answer> 
No extra text outside the two tags. 
Keep the response under 50 words. 
Example: 
<think>Row 1 has one empty cell -> place 1. Column 2 then needs 2.
</think>
<answer>1,3,1 || 3,2,2</answer>
\end{promptbox}

\begin{promptbox}{Sudoku Prompt: Observation then prediction}
You are solving 4x4 Sudoku. 
Fill empty cells with digits 1-4.
Use a 1-indexed grid (rows/cols start at 1). 
A move is exactly: row,col,value (three integers). 
In one turn you may output multiple moves, separated by ||. 
Only propose moves that keep the row, column, and 2x2 subgrid valid. Always output EXACTLY as: 
<think>[brief reasoning]</think> 
<answer>[r,c,v || r,c,v ...]</answer> 
No extra text outside the two tags. 
Keep the response under 50 words. 

An example output: 
<think> 
<observation>
| . . 1 4 | 1 4 . 3 | 4 2 . . | . 1 4 2 
Empty positions to be filled are at (1,1), (1,2), (2,3), (3,3), (3,4), (4,1)
</observation> 
<prediction>
| 2 3 1 4 | 1 4 2 3 | 4 2 3 1 | . 1 4 2
Empty positions to be filled are at (4,1)
</prediction> 
</think> 
<answer> 1,1,2 || 1,2,3 || 2,3,2 || 3,3,3 || 3,4,1 </answer>.
\end{promptbox}

%% file: sections/appendix-expsetup.tex
\section{Experimental Setup}

These text-based game environments we evaluate are highly controllable; we adjust task difficulty by varying grid size to match model capacity. For Sokoban, we fix a $6\times6$ grid for all models. For FrozenLake, we use $4\times4$ for smaller models (Qwen2.5-0.5B-Instruct, Qwen2.5-1.5B-Instruct and LLaMA-3.2-1B-Instruct) and $6\times6$ for Qwen2.5-3B to avoid ceiling effects observed with the $4\times4$ setting. For Sudoku, we use a $4\times4$ grid with six empty cells to be filled.

Experiments run on up to 8 $\times$ 80G NVIDIA H100. For training, we instruction-tune on 1280 samples using AdamW with learning rate $1\times10^{-4}$. We apply PPO initialized from the SFT checkpoint, using a learning rate of $1\times10^{-6}$ for actor and $1\times10^{-5}$ for critic. It takes about 12 hours on single GPU to train Qwen2.5-1.5B-Instruct for 1,000 steps.
For evaluation, we report decoding-only results with temperature $1.0$, top-$p$ $1.0$, and max new tokens $400$.
\camera{Our code, configuration files, and trained checkpoints are publicly available at \url{https://github.com/shiqichen17/SPA}.}

%% file: sections/appendix-fullres.tex
\input{Tables/fullres}

%% file: Tables/fullres.tex
\begin{table*}[t]
\centering
\begingroup
\setlength{\tabcolsep}{6pt}
\renewcommand{\arraystretch}{1.0}
\caption{Results on Sokoban, FrozenLake, and Sudoku for different models. (Metrics in $\times 10^{-2}$).}
\scalebox{0.95}{
\begin{tabular}{>{\raggedright\arraybackslash}p{3.8cm}p{1.2cm}p{1.2cm}p{1.2cm}p{1.2cm}p{1.2cm}p{1.2cm}}
\toprule
\multirow{2}{*}{\textbf{Model}} & \multicolumn{2}{c}{\textbf{Sokoban}} & \multicolumn{2}{c}{\textbf{FrozenLake}} & \multicolumn{2}{c}{\textbf{Sudoku}} \\
\cmidrule(lr){2-7}
& \textbf{Pass@1} & \textbf{Pass@8} & \textbf{Pass@1} & \textbf{Pass@8} & \textbf{Pass@1} & \textbf{Pass@8} \\
\midrule
Qwen2.5-0.5B-Instruct         & 5.5  & 25.8 & 8.1  & 32.8 & 0.0  & 0.0  \\
PPO                           & 16.9 & 35.9 & 22.8 & 30.1 & 0.0  & 0.0  \\
+State Estimation             & 5.2  & 25.4 & 6.7  & 28.5 & 0.1  & 0.8  \\
+State Estimation RL          & 20.4 & 38.7 & 24.2 & 26.6 & 4.5  & 14.1 \\
+~\texttt{SPA} SFT            & 1.8  & 11.7 & 3.1  & 15.6 & 0.1  & 0.8  \\
+~\texttt{SPA}         &
\cellcolor{MyCellShade}36.7 & \cellcolor{MyCellShade}45.3 &
\cellcolor{MyCellShade}46.9 & \cellcolor{MyCellShade}65.6 &
\cellcolor{MyCellShade}18.2 & \cellcolor{MyCellShade}43.8 \\
+VAGEN                        & 33.3 & 44.9 & 25.8  & 40.6  & 0.1  & 0.8  \\
\midrule
Qwen2.5-1.5B-Instruct         & 16.3 & 47.3 & 17.2 & 49.2 & 1.6  & 11.3 \\
PPO                           & 25.6 & 34.0 & 22.1 & 30.7 & 0.0  &  0.0 \\
+State Estimation             & 15.7 & 43.8 & 18.6 & 51.2 & 0.3  &  2.3 \\
+State Estimation RL          & 52.7 & 53.9 & 27.6 & 34.8 & 39.1 & 72.3 \\
+~\texttt{SPA} SFT             & 8.3  & 36.7 &  9.8 & 43.8 & 3.5  & 24.2 \\
+~\texttt{SPA}         &
\cellcolor{MyCellShade}59.8 & \cellcolor{MyCellShade}69.5 &
\cellcolor{MyCellShade}70.9 & \cellcolor{MyCellShade}75.0 &
\cellcolor{MyCellShade}59.6 & \cellcolor{MyCellShade}94.9 \\
+VAGEN                        &  44.5  &  50.0  &  37.7   &  43.0  &  0.0   &  0.0   \\
\midrule
Qwen2.5-3B                    & 12.5 & 35.9 &  6.9 & 26.9 & 0.0  &  0.0 \\
PPO                           & 31.4 & 35.5 &  7.6 & 16.0 & 0.0  &  0.0 \\
+State Estimation             &  9.9 & 39.3 &  6.9 & 26.6 & 0.1  &  0.8 \\
+State Estimation RL          & 26.2 & 27.7 & 24.7 & 35.5 & 24.1 & 26.2 \\
+~\texttt{SPA} SFT             &  6.0 & 29.7 &  1.6 & 10.5 & 0.3  &  2.7 \\
+~\texttt{SPA}         &
\cellcolor{MyCellShade}49.7 & \cellcolor{MyCellShade}58.2 &
\cellcolor{MyCellShade}41.3 & \cellcolor{MyCellShade}46.1 &
\cellcolor{MyCellShade}69.9 & \cellcolor{MyCellShade}89.8 \\
+VAGEN                        &  31.9  &  43.0   &  30.7   &  34.0   &  0.0  &  0.0   \\

\midrule
LLaMA3.2-1B-Instruct          &  8.3 & 33.2 & 10.2 & 41.0 &  0.0 &  0.0 \\
PPO                           & 21.2 & 39.8 & 10.8 & 24.6 &  0.1 &  1.2 \\
+State Estimation             &  9.6 & 33.6 & 10.9 & 43.0 &  0.1 &  0.8 \\
+State Estimation RL          & 31.6 & 44.1 & 19.3 & 29.7 &  45.0&  71.1\\
+~\texttt{SPA} SFT             &  6.3 & 27.7 & 11.4 & 38.3 &  1.3 &  9.8 \\
+~\texttt{SPA}         &
\cellcolor{MyCellShade}53.0 & \cellcolor{MyCellShade}68.0 &
\cellcolor{MyCellShade}64.8 & \cellcolor{MyCellShade}71.1 &
\cellcolor{MyCellShade}81.3 & \cellcolor{MyCellShade} 100 \\
+VAGEN                        & 47.4 & 50.8 & 24.5 & 29.3 &  0.1  & 0.8 \\
\midrule
GPT-OSS-20B                   & 45.8 & 84.4 & 68.8 & 100  &  61.8 &  100   \\
+State Estimation             & 55.1 & 89.1 & 73.3 & 100  &  68.1  &  100   \\
\bottomrule
\end{tabular}
\label{tab:fullres}

}
\endgroup
\end{table*}
Table~\ref{tab:fullres} reports the complete results. We also include the performance before RL training (step = 0) for both settings: (i) State‑Estimation PPO, evaluated with the State‑Estimation prompt, and (ii) \texttt{SPA} immediately after the world‑modeling SFT stage. At step 0, the State‑Estimation prompt yields better performance than baseline, whereas the additional world‑modeling SFT slightly lowers accuracy. However, after subsequent RL training, \texttt{SPA} quickly recovers and ultimately surpasses State‑Estimation PPO across tasks, indicating that the structure learned during SFT provides a transferable inductive bias that RL can exploit for better generalization.

%% file: sections/appendix-analyses.tex
\section{Confound-Controlled Ablations and Direct Exploration Metrics}
\label{appdx:analyses}

\input{Tables/plan_pred}
\input{Tables/coverage}
\begin{camerablock}
\subsection{The prediction slot is not redundant with the plan}
\label{appdx:planpred}
Free-form planning text inside the reasoning trace could, in principle, account for the gains attributed to next-state prediction. To isolate the two without confounds, we compare three aligned variants on Sokoban (Qwen2.5-1.5B-Instruct, Pass@1/Pass@8, no state estimation): (i) \emph{Vanilla RL}, whose reasoning contains a plan but no prediction; (ii) a \emph{prediction-only} \texttt{SPA} variant with a grounded next-state span and action but no separate plan; and (iii) \texttt{SPA} w/o state estimation, which keeps both plan and prediction. As shown in Table~\ref{tab:planpred}, keeping the plan but dropping prediction (25.6/34.0) falls well below plan+prediction (36.3/40.6), so next-state prediction adds a substantial, separable gain even when a plan is present. Conversely, dropping the plan (28.9/37.3) also falls below plan+prediction yet still exceeds the plan-only baseline. The two slots are therefore complementary, and the takeaway does not reduce to ``grounded observation plus chain-of-thought is enough''.

\subsection{The policy refines, rather than ignores, its predictions during RL}
\label{appdx:trace}
If the model simply learned to ignore the \tokennexts{} slot, its prediction accuracy would stagnate or decay during RL. We track the next-state matching accuracy of on-policy rollouts over the full \texttt{SPA} Sokoban run and find that it rises from 31.9\% right after SFT to 44.4\% by step 100 and 46.6\% averaged over steps 301--500, then largely plateaus while task success keeps climbing (train Pass@1 rises from 12.9 to 30.8 over the same span, reaching 59.8 at final validation). The overall gain therefore reflects SFT initialization, early transition refinement, and later RL task optimization together---consistent with \texttt{SPA} acting as a grounded initialization recipe rather than an accurate online simulator.

\subsection{Beyond Pass@k: measuring useful exploration directly}
\label{appdx:spcr}
Pass@k is an indirect proxy for exploration: a change in Pass@k can reflect rollout correlation, sampling temperature, or solution diversity rather than coverage of meaningful states. We therefore complement it with post-hoc trajectory metrics. Let $\mathcal{U}(\pi)$ denote the set of unique states visited by rollouts of policy $\pi$, and $\mathcal{U}^{+}(\pi)\subseteq\mathcal{U}(\pi)$ those states lying on at least one successful trajectory; we define the \emph{successful-path coverage ratio} (SPCR) $\rho(\pi)=|\mathcal{U}^{+}(\pi)|/|\mathcal{U}(\pi)|$. This makes our view of the exploration--exploitation tradeoff explicit: the terminal objective remains the expected return $J(\pi)$, exploration quantities such as coverage $|\mathcal{U}(\pi)|$ and action diversity are instrumental rather than terminal, and coverage helps only insofar as it concentrates on success-relevant states---high $\rho(\pi)$ at high $J(\pi)$, not maximal $|\mathcal{U}(\pi)|$.

Table~\ref{tab:coverage} compares weaker (vanilla) runs against stronger grounded (\texttt{SPA}) runs: the grounded runs visit \emph{fewer} total states ($40\to25$ on Sokoban, $32\to23$ on FrozenLake) yet cover far more success-relevant states and exhibit higher action diversity (mean pairwise edit distance between sampled action sequences). The trend is robust: among the 12 grounded FrozenLake final-step runs we located, 10/12 outperform the raw baseline in final success and 10/12 in SPCR. The precise claim is therefore not ``more exploration'' in a naive coverage sense, but \emph{more useful exploration}.
\end{camerablock}

%% file: Tables/plan_pred.tex
\begin{table}[t]
\centering
\small
\setlength{\tabcolsep}{4pt}
\camera{%
\begin{tabular}{lcc}
\toprule
Method (Sokoban) & Pass@1 & Pass@8\\
\midrule
Vanilla RL (plan, no prediction) & 25.6 & 34.0 \\
SPA prediction-only (no plan) & 28.9 & 37.3 \\
SPA w/o State Est.\ (plan + prediction) & 36.3 & 40.6 \\
\bottomrule
\end{tabular}}
\caption{\camera{Confound-free decomposition of the reasoning slots on Sokoban (Qwen2.5-1.5B-Instruct, no state estimation). The prediction span is not redundant with free-form planning, and neither component alone recovers the full combination.}}
\label{tab:planpred}
\end{table}

%% file: Tables/coverage.tex
\begin{table}[t]
\centering
\small
\setlength{\tabcolsep}{4pt}
\camera{%
\begin{tabular}{lccc}
\toprule
Environment & Success & SPCR $\rho$ & Action div.\\
\midrule
Sokoban & 0.30$\,\to\,$0.65 & 0.125$\,\to\,$0.480 & 0.561$\,\to\,$0.681 \\
FrozenLake & 0.30$\,\to\,$0.75 & 0.281$\,\to\,$0.783 & 0.586$\,\to\,$0.699 \\
\bottomrule
\end{tabular}}
\caption{\camera{Direct trajectory metrics for weaker (vanilla) $\to$ stronger grounded (\texttt{SPA}) runs: success rate, successful-path coverage ratio $\rho$, and action diversity (mean pairwise edit distance between sampled action sequences).}}
\label{tab:coverage}
\end{table}

%% file: sections/appendix-llm.tex
During the preparation of this paper, we used large language models (LLMs) such as ChatGPT for assistance in polishing the writing and improving the clarity of figures. LLMs were not used for research ideation, experimental design, or the production of experimental results. All conceptual contributions, analyses, and conclusions are solely those of the authors. The authors take full responsibility for the content of this paper.

%% file: sections/appendix-alfworld.tex
To evaluate whether \texttt{SPA} can generalize to more realistic settings, we also evaluate \texttt{SPA} on Alfworld using the Qwen2.5-0.5B-Instruct model.
We choose a smaller 0.5B model as we observe that the performance quickly saturates for the 1.5B model (in Figure~\ref{fig:teaser}). 
The results are in Table~\ref{tab:0.5b-alfworld-passk-results}. 
We observe that \texttt{SPA} delivers consistent gains on both the Pass@1 and Pass@k metrics. 
This demonstrates that our method can be effective beyond text games. 

\begin{table}[t]
    \centering
    \caption{ Performance of Qwen2.5-0.5B-Instruct on Alfworld.}
    \label{tab:0.5b-alfworld-passk-results}
    \begin{tabular}{lcc}
        \toprule
         Method &  Pass@1 &  Pass@8 \\
        \midrule
         Baseline &  30.4 &  76.5 \\
         SPA &  52.1 &  93.3 \\
        \bottomrule
    \end{tabular}
\end{table}

%% file: sections/appendix-grpo.tex
To evaluate whether \texttt{SPA} can also work on other RL algorithms, we also evaluate \texttt{SPA} on GRPO using the Qwen2.5-1.5B-Instruct model. 
The results are in Table~\ref{tab:grpo}. 
We observe that \texttt{SPA} delivers consistent gains on both the Pass@1 and Pass@k metrics. 
This demonstrates that our method can be effective across the RL algorithms. 

\begin{table}[t]
    \centering
    \small
    \setlength{\tabcolsep}{4pt}
    \caption{Performance of Qwen2.5-1.5B-Instruct on Sokoban and FrozenLake (GRPO).}
    \label{tab:grpo}
    \begin{tabular}{lcccc}
        \toprule
        \multirow{2}{*}{Method (GRPO)} & \multicolumn{2}{c}{Sokoban} & \multicolumn{2}{c}{FrozenLake} \\
        \cmidrule(lr){2-3}\cmidrule(lr){4-5}
        & Pass@1 & Pass@8 & Pass@1 & Pass@8 \\
        \midrule
        Baseline RL & 0.0 & 0.0 & 0.0 & 0.0 \\
        State Estimation RL & 40.5 & 45.9 & 31.3 & 34.8 \\
        + SPA & \textbf{42.2} & \textbf{52.5} & \textbf{74.1} & \textbf{80.6} \\
        \textbf{$\Delta$ (Improvement)} & \textbf{+1.7 $\uparrow$} & \textbf{+6.6 $\uparrow$} & \textbf{+42.8 $\uparrow$} & \textbf{+45.8 $\uparrow$} \\
        \bottomrule
    \end{tabular}
\end{table}